\documentclass[referee,sn-mathphys-num]{sn-jnl}

\usepackage{graphicx}%
\usepackage{multirow}%
\usepackage{amsmath,amssymb,amsfonts}%
\usepackage{amsthm}%
\usepackage{mathrsfs}%
\usepackage[title]{appendix}%
\usepackage{xcolor}%
\usepackage{textcomp}%
\usepackage{manyfoot}%
\usepackage{booktabs}%
\usepackage{algorithm}%
\usepackage{algorithmicx}%
\usepackage{algpseudocode}%
\usepackage{listings}%

\theoremstyle{thmstyleone}%
\theoremstyle{thmstyletwo}%

\theoremstyle{thmstylethree}%

\begin{document}

\title[Article Title]{GTAGCN: Generalized Topology Adaptive Graph Convolutional Networks}


\author[2]{\fnm{Sukhdeep} \sur{Singh}}\email{sukha13@ymail.com}
\author*[1]{\fnm{Anuj} \sur{Sharma} \footnote{Corresponding Author}}\email{anujs@pu.ac.in}


\author[3]{\fnm{Vinod Kumar} \sur{Chauhan} \footnote{Principal Author}}\email{vinod.kumar@eng.ox.ac.uk}
\affil*[1]{\orgdiv{Department of Computer Science and Applications}, \orgname{Panjab University}, \orgaddress{\city{Chandigarh}, \country{India}}}

\affil[2]{\orgdiv{Department of Computer Science}, \orgname{DM College (aff. to Panjab University, Chandigarh)}, \orgaddress{\city{Moga}, \state{Punjab}, \country{India}}}

\affil[3]{\orgdiv{Department of Engineering Science}, \orgname{University of Oxford}, \orgaddress{\city{Oxford}, \country{UK}}}

\abstract{Graph Neural Networks (GNN) has emerged as a popular and standard approach for learning from graph-structured data. The literature on GNN highlights the potential of this evolving research area and its widespread adoption in real-life applications. However, most of the approaches are either new in concept or derived from specific techniques. Therefore, the potential of more than one approach in hybrid form has not been studied extensively, which can be well utilized for sequenced data or static data together. We derive a hybrid approach based on two established techniques as generalized aggregation networks and topology adaptive graph convolution networks that solve our purpose to apply on both types of sequenced and static nature of data, effectively. The proposed method applies to both node and graph classification. Our empirical analysis reveals that the results are at par with literature results and better for handwritten strokes as sequenced data, where graph structures have not been explored.

}




\maketitle

\section{Introduction}
\label{intro}
Graph Neural Networks (GNN) have emerged as a suitable choice for non-euclidean data or irregular data, where Graph Convolutional Networks (GCN) are able to work efficiently for these complex irregular data. The GCN advancements provide meaningful and valuable relationships between individuals in social networks \cite{Easley2010} or chemical interactions \cite{Stokes2020} or recommending engines \cite{Monti2017} or drug discovery \cite{Zitnik2017,Wale2008} and similar problems. This has resulted in the availability of graph structured data and techniques associated with GNN that could handle such datasets. This area of deep learning introduced many successful algorithms which were missing before GNN implementation. The graphs and their rich Mathematical properties were systematically used in GNN which inherits the properties message passing among nodes, aggregation and graph kernels \cite{Gilmer2017,Battaglia2018, Xu2019b}. Notably, most of the GCN were either spatial or spectral in nature \cite{Bruna2014,Levie2017}, where data based on graph Fourier transformation is spectral in nature and spatial GCN consider use spatial features \cite{Kipf2016,Monti2017}. The feature space can be systematically applied to spectral and spatial features which include feature subspaces flattening and structural principal components to expand feature space \cite{sunicml23}.
\par The GCN have been studied extensively in the recent past and work was mainly focused on message passing \cite{Gilmer2017}. Most of the developed techniques follow recursive neighborhood aggregation, where a node aggregates messages from its neighboring nodes and updates itself. The updated nodes are permutation invariant and aggregation functions mainly include mean \cite{Kipf2016}, max \cite{Hamilton2017} and powermean \cite{Xu2019b}. Recent work has been carried out in the direction of permutation invariant using relational pooling where explicit labels are assigned to nodes as additional features \cite{zhouicml23}. Also, graph convolution operations multiplication by the graph adjacency matrix appeared to be a suitable choice in the previous study \cite{Sandryhaila2013} \cite{tagcn2018}. Further, polynomials of graph adjacency matrix opened new directions to enhance GNN for time or image based signal processing systems. This has motivated us to solve the following question for GNN,\\
``What could be the impact of aggregation functions and polynomials of graph adjacency matrix together to sequenced or static nature of data?"
\par We have answered this question by combining two properties as aggregation function and polynomials of graph adjacency matrix, and tested for the pattern related problems, where time based data and static images based data were both evaluated. The time based data has been taken as online handwriting patterns \cite{ijdar2017} and scanned images are used as image data \cite{mnist}. In addition, we have used other benchmarked data, which are different from handwritten pattern data. 
\par The GNN could work for supervised and unsupervised problems both. The supervised learning includes graph kernels level and message passing for node, link, and graph level transductive tasks. The unsupervised learning includes shallow embeddings and message passing tasks as in supervised cases \cite{matthias2022}. It has been noticed that mostly GNNs are used for supervised learning \cite{mtap2023}. The present study is in the direction of supervised learning which include model building based on training data and its evaluation subject to test data. Most of the graph related problems are NP-hard in nature such as the evaluation of the best connection of nodes combination or the best path in the graph. The time based data has been used in sequential order that is able to reduce NP-hard complexities to an extent. For other data, we have used the graphs as provided by the sources. 
\par In this paper, we have proposed Generalized Topology Adaptive Graph Convolutional Networks (GTAGCN) based on two successful techniques \cite{gen2020}. This scheme allows us to derive different variants of the proposed model. This technique focuses on the feasibility of computation for message passing task and maintaining the balance of the network. The graph filters refer to polynomials of the adjacency matrix in graph signal processing \cite{tagcn2018} and generalized aggregation functions are differentiable that help to learn better. Our main contributions are summarized as follows:
\begin{itemize}
	\item The proposed GTAGCN combines two established techniques as generalized aggregation networks and topology adaptive GCN systematically that results in the smooth working of the proposed GTAGCN GCN.
	\item The results are best reported for time based data as online handwriting patterns and at par or close to other image based data.
	\item The proposed GTAGCN accepts K-localized filters as happen in TAG GCN to extract local features on a set of sizes from 1 to K receptive fields.
	\item In addition, generalized aggregation networks use of MLP and RELU are used in GTAGCN.
\end{itemize}
The paper has been organized as follows. Section 2 includes related work and section 3 discusses the necessary introduction of GNN including notations. Section 4 explains the theoretical framework and proposed algorithm. Section 5 presents the results and discussion. The last section 6 concludes this paper with findings.

\section{Notations and Basics}
\begin{figure}[h]
\begin{center}
\includegraphics[width=10cm]{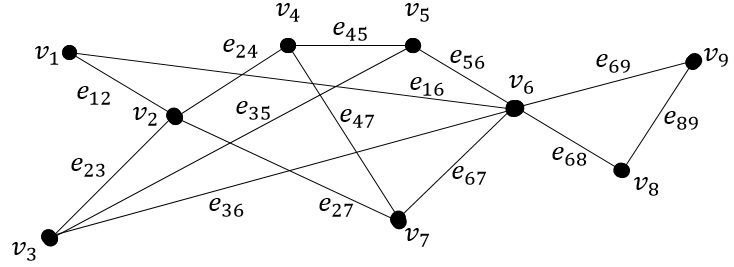}
\caption{Nodes and edges in a graph}
\label{gentag-fig1}
\end{center}
\end{figure}

A graph $\mathcal{G}$ is defined as $\mathcal{G=(V,E)}$, where $\mathcal{V}=\{v_1,v_2,...,v_N\}$ and $\mathcal{E} \subseteq \mathcal{V} \times \mathcal{V}$ are the set of vertices and edges, respectively. An edge $e$ from node $u \in \mathcal{V}$ to node $v \in \mathcal{V}$ is represented as $e=(u,v) \in \mathcal{E}$. The nodes and edges in a graph are depicted in figure \ref{gentag-fig1}.
An adjacency matrix is $A \in \mathbb{R}^{\mathcal{|V|} \times \mathcal{|V|}}$.

\par One of the GNN fundamental components is nodes that enable network to learn structural representation. The edges are connections of nodes and share rich Mathematical properties as one-to-one, one-to-many and many-to-many relationships. The encoded features of graphs are important structure information of edges. The GNN's other component is graph convolutional layer that plays a major role in learning network from graph structure and data \cite{layer1000s2021}. The node feature vector with enriched characteristics is part of the graph convolutional layer. The neighboring nodes feature vector aggregate to current node information using graph convolution operator function which is part of graph convolutional layer \cite{dlsurvey2022}. The convolution operators are special operators different from traditional operators and it includes learning of feature weights. This convolutional operation is performed on neighboring nodes and their associated features. This working of GNN operators uses non-linear activation functions and results feature vector passed to the next layers. This happen in an iterative mode for graph convolutional layers that help the network to learn weights which is an effective representation of graph structured data able to capture important information of graphs \cite{gnncomputation2008}. The activation functions are used to introduce non-linearity to layers output that further contribute in learning complex patterns of data \cite{activation2018}. The nodes in layers output applied by activation functions and aggregated with neighboring node features. Common activation functions are rectified linear unit (ReLU ($max(0,x)$)), sigmoid ($\frac{1}{1+e^{-x}}$), and hyperbolic tangent ($\frac{1-e^{-2x}}{1+e^{-2x}}$) functions. Pooling and aggregation are used in GNN to reduce the dimensionality of the feature space and enable the network to handle graphs of different sizes. The pooling involves aggregating information from multiple nodes or subgraphs into a single representation \cite{pooling2022, aggregation2020}. This mainly includes max pooling, mean pooling, or sum pooling. 

\par The neural message passing is one of the key features of GNN which include vector messages are exchanged between nodes and update the node \cite{Gilmer2017}. The message passing can be understood as follows,
\begin{equation}
	\begin{split}
		\mathbf{h}_{u}^{(l+1)} = \mathrm{UPDATE}^{(l)}(\mathbf{h}_{u}^{(l)},\\ \mathrm{AGGREGATE}^{(l)}({\mathbf{h}_{v}^{(l)}}, \forall v \in \mathcal{N}(u)))
	\end{split}
\end{equation}
Here, $\mathrm{AGGREGATE}^{(l)}({\mathbf{h}_{v}^{(l)}}, \forall v \in \mathcal{N}(u))$ is the aggregated message and expressed as $m_{\mathcal{N}(u)}^{(l)}$. 
\begin{figure}[h]
\begin{center}
\includegraphics[width=10cm]{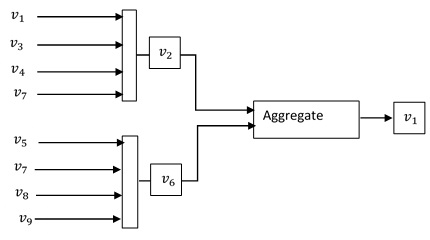}
\caption{Single node aggregates messages overview from its local neighborhood.}
\label{gentag-fig2}
\end{center}
\end{figure}
The $\mathbf{h}_{v}^{(l)}$ refers to vertex features.  This results in final output layer as, $z_{u}=h_{u}^{L}, \forall u \in \mathcal{V}$. This abstract form can be expressed as basic GNN message passing,
\begin{equation}
	\begin{split}
		\mathbf{h}_{u}^{(l+1)} = \sigma(\mathbf{W}_{self}^{(l+1)} \mathbf{h}_{u}^{(l)}+\mathbf{W}_{neigh}^{(l+1)} \sum_{v\in \mathcal{N}(u)}\mathbf{h}_{u}^{(l)}+\mathbf{b}^{(l)})
	\end{split}
\end{equation}
where, $\mathbf{W}_{self}^{(l+1)}$, $\mathbf{W}_{neigh}^{(l+1)}$ $\in$ $\mathbb{R}^{d(l) \times d(l-1)}$ are trainable parameter matrices and $\sigma$ is elementwise non-linearity as $tanh$ or $\mathrm{ReLU}$ \cite{william2020}. This phenomenon can be understood with figure \ref{gentag-fig2}. Figure \ref{gentag-fig2} presents the working of message passing for the node $v_1$ of figure \ref{gentag-fig1}. The node $v_1$ is directly connected to nodes $v_2$ and $v_6$. The model aggregates messages from $v_1$ local graph neighbors as $v_2$ and $v_3$, and in turn, the messages coming from these neighbors ($v_2, v_6$) are based on information aggregated from their respective neighborhoods ($v_2$ depends on $v_1, v_3, v_4, v_7$ and $v_6$ depends on $v_5, v_7, v_8, v_9$), and so on. Therefore, figure \ref{gentag-fig2} is GNN computation graph of tree structure nature and two-layer version of message passing model. One of the baselines of GNN model is GCN \cite{Kipf2016} which defines the message passing function as,
\begin{equation}
	\begin{split}
		\mathbf{h}_{u}^{(l)} = \sigma \left(\mathbf{W}^{(l)} \sum_{v\in \mathcal{N}(u)\cup \{v\}} \dfrac{\mathbf{h}_{v}}{\sqrt{|\mathcal{N}(u)||\mathcal{N}(v)|}} \right)
	\end{split}
\end{equation}
and the basic GCN layer is defined as,
\begin{equation}
	\begin{split}
		\mathbf{H}^{(l)} = \sigma \left(\hat{A}\mathbf{H}^{(l-1)}\mathbf{W}^{l} \right)
	\end{split}
\end{equation}
where, $\hat{A}=D^{-1/2}AD^{-1/2}$ and $D$ is the diagonal degree matrix of $\hat{A}$.

\section{Related Work}
\label{relatedwork}
The GNN have made significant advancements in the recent past \cite{mtap2023}. The present study presents a new GNN operator based on literature work. Selected graph convolution operators are discussed here. One of the early and popular works in this direction is Graph Convolutional Network Convolution \cite{gcnconv2017} which represents working as,
\begin{equation}
\mathbf{H}^{(l+1)} = {\mathbf{D}}^{-\frac{1}{2}} {\mathbf{A}} {\mathbf{D}}^{-\frac{1}{2}} \mathbf{H}^{(l)} \mathbf{W}^{(l)}   
\end{equation}
The other operator that follows spectral graph theory and efficient numerical schemes to design fast localized convolutional filters on graphs is Chebyshev graph convolutional \cite{chebconv2016}. It can be expressed as,
\begin{equation}
\mathbf{H}^{(l+1)} = \sum_{k=0}^{K-1} \mathbf{\Theta}^{(l)}_k \mathbf{T}_k(\tilde{\mathbf{L}}) \mathbf{H}^{(l)} \mathbf{W}^{(l)}
\end{equation}
where, $\mathbf{\Theta}^{(l)}_{k} \in \mathbb{R}^{d^{(l)} \times d^{(l+1)}}$ is the $k^{th}$ learnable Chebyshev filter at layer $l$, $\tilde{\mathbf{L}} = 2\mathbf{L}/\lambda_{max}-\mathbf{I}_N$ is the normalized graph Laplacian with eigenvalues scaled to the range $[-1, 1]$ and $\mathbf{L} = \mathbf{I}_N - \mathbf{A}$ refers to Laplacian matrix. $\mathbf{T}_k(\tilde{\mathbf{L}})$ is the $k^{th}$ Chebyshev polynomial of degree $k$ evaluated at $\tilde{\mathbf{L}}$.
To address and efficiently generate node embeddings in unseen data is Sample and aggregate convolutional network Convolution \cite{sageconv2017}, which can be expressed as,
\begin{equation}
\mathbf{H}^{(l+1)}_i = \sigma \left( \mathbf{W}^{(l)} \cdot \text{CONCAT} \left( \mathbf{H}^{(l)}_i, \text{MEAN}_{j \in \mathcal{N}(i)} \{\mathbf{H}^{(l)}_j\} \right) \right)
\end{equation}
The gated graph convolution \cite{gatedgraphconv2016} based on Gated Recurrent Unit includes,
\begin{equation}
\mathbf{H}^{(l+1)}= \sigma \left(j(h^{k,T},\mathbf{H}^{(l)}) \right)
\end{equation}
The Graph Attention Network v2 Convolution (GATv2Conv) \cite{gatv2conv2022} is another extension of gated graph convolution using message passing, attention and update. This can be written as,
\begin{equation}
e_{ij}^{(l)}= \text{LeakyReLU} \left({\mathbf{a}}^{(l)T} [\mathbf{W}^{(l)}\mathbf{H}_i^{(l)}, \mathbf{W}^{(l)}\mathbf{H}_j^{(l)}] \right)
\end{equation}
\begin{equation}
\alpha_{ij}^{(l)}= \frac{\text{exp}(\text{LeakyReLU}(\mathbf{e}_{ij}^{(l)}))}{\sum_{k \in \mathcal{N}_i} \text{exp}(\text{LeakyReLU}(\mathbf{e}_{ik}^{(l)}))}
\end{equation}
\begin{equation}
\mathbf{H}_i^{(l+1)}= \sigma \left( \mathbf{W}^{(l)} \left( \sum_{j \in \mathcal{N}_i} \alpha_{ij}^{(l)} \mathbf{H}_j^{(l)} + \mathbf{H}_i^{(l)} \right) \right)
\end{equation}
The Unified Message Passaging Model \cite{TransformerConv2021} includes feature and label propagation at both training and inference time.
\begin{equation}
v^{(l)}_{c;j} = W^{(l)}_{c;v} h^{(l)}_{j} + b^{(l)}_{c;v}
\end{equation}
\begin{equation}
\hat{h}^{(l+1)}_i = \||_{c=1}^{C} \sum_{j \in N(i)} \mathbf{\alpha}^{(l)}_{c;ij} \left(v^{(l)}_{c;j} + e_{c,{ij}}\right)
\end{equation}
where, the $||$ is the concatenation operation for $C$ head attention. 
The Autoregressive Moving Average Convolution \cite{armaconv2019} was based on the autoregressive moving average as,
\begin{equation}
\mathbf{H}^{(l+1)} = \sigma \left( \mathbf{\hat{L}}
\mathbf{H}^{(l)} \mathbf{W} + \mathbf{H}^{(0)} \mathbf{V} \right),
\end{equation}
where, $\mathbf{\hat{L}} = \mathbf{I} - \mathbf{L} = \mathbf{D}^{-1/2} \mathbf{A} \mathbf{D}^{-1/2}$. The Gaussian Mixture Model Convolution \cite{gmmconv} includes the output feature vector of each node computed as a weighted sum of the feature vectors of the Gaussian mixture components, with probability matrix $\mathbf{P}_{i,j}^{(l)}$. It can be expressed as,
\begin{equation}
\mathbf{H}^{(l+1)} = \sum_{i=1}^{M} \sum_{j=1}^{N} \mathbf{P}_{i,j}^{(l)} \cdot \sum_{k=1}^{K} \mathbf{H}_{j,k}^{(l)} \cdot \mathbf{\Theta}_{i,k}^{(l)}
\end{equation}
\begin{equation}
\mathbf{P}_{i,j}^{(l)} = \mathrm{softmax}\left( \frac{\mathrm{exp}\left(-\frac{1}{2}\cdot \left(\mathbf{x}_{i} - \mathbf{y}_{j} \right)^{T} \cdot \mathbf{A} \cdot \left(\mathbf{x}_{i} - \mathbf{y}_{j} \right)\right)}{\sum_{j'=1}^{N} \mathrm{exp}\left(-\frac{1}{2}\cdot \left(\mathbf{x}_{i} - \mathbf{y}_{j'} \right)^{T} \cdot \mathbf{A} \cdot \left(\mathbf{x}_{i} - \mathbf{y}_{j'} \right)\right)} \right)
\end{equation}
where, $\mathbf{P}_{i,j}^{(l)}$ is a probability matrix that assigns each node $i$ to a set of $K$ Gaussian mixture components centred at nodes $j$, $\mathbf{H}_{j,k}^{(l)}$ is the feature vector of the $k$-th component centred at node $j$, $\mathbf{\Theta}_{i,k}^{(l)}$ is the weight matrix of the $k$-th component assigned to node $i$, $\mathbf{x}_{i}$ is the feature vector of node $i$, $\mathbf{y}_{j}$ is the feature vector of node $j$, $\mathbf{A}$ is a learnable affinity matrix that controls the similarity between nodes, and $\mathrm{softmax}$ is a softmax function that ensures that the weights assigned to each Gaussian mixture component sum to 1.
The B-spline functions based work reported in Spline Convolution \cite{splineconv2018} which includes,
\begin{equation}
(\mathbf{f \star g})(i)=\frac{1}{\mathcal{N}(i)} \sum_{i=1}^{M_m} \sum_{j \in \mathcal{N}(i)} f_l(j) \cdot g_l(u(i,j))
\end{equation}
The other advancements made as hypergraph convolution \cite{hypergraphconv2020} is a generalization of graph convolution to hypergraphs and each node is connected to a fixed number of neighbors. It can be expressed as, 
\begin{equation}
H^{(l+1)} = \left(D^{-\frac{1}{2}} A W B^{-1} A^T D^{-\frac{1}{2}}\right) H^{(l)} P
\end{equation}
and its attention as,
\begin{equation}
H_{ij} = \sum_{k \in N_i} \frac{\exp\left(-\frac{1}{2}\|\mathbf{x}_i^P - \mathbf{x}_j^P\|^2\right)}{\sum_{k' \in N_i} \exp\left(-\frac{1}{2}\|\mathbf{x}_i^P - \mathbf{x}_k^P\|^2\right)},
\end{equation}
where $\|\mathbf{x}_i^P - \mathbf{x}_j^P\|$ denotes the Euclidean distance between the transformed feature vectors of nodes $i$ and $j$. The Extended Convolution \cite{xconv2018} operator was suitable for irregular and unstructured points cloud data. It can be expressed as,
\begin{equation}
F_p = X\text{Conv}(K, p, P, F) = \text{Conv}(K, \text{MLP}(P - p) \, \times \, [\text{MLP}(P - p), F])
\end{equation}
$\text{MLP}$ is a multi-layer perceptron that takes as input the maximum feature vector among the neighboring points of point $i$. The $F$ is trainable convolution kernels and $P$ refers to features. Many GNN operators were reported in the literature.
\par In addition to above mentioned literature work, our work is closely related to the following two techniques.
\par \textbf{Topology Adaptive Graph Convolutional Networks (TAGCN)}
explores a $K$-localized filter for graph convolution in the vertex domain that helps to extract local features up to size $K$ \cite{tagcn2018}. The TAGCN includes $G_{c,f}^{(l)}\in \mathbb{R}^{N_l \times N_l}$ denote the $f$-th graph filter and $l$ is $l$-th hidden layer. The $G_{c,f}^{(l)}x_{c}^{(l)}$ is graph convolution and $f$-th output filter followed by ReLU function is,
\begin{equation}
	\begin{split}
		\mathbf{y}_{f}^{(l)} = \sum_{c=1}^{C_l} G_{c,f}^{(l)}x_{c}^{(l)} + b_f \mathbf{I}_{N_l}
	\end{split}
\end{equation}
where, $b_f$ is a learnable bias and $\mathbf{I}_{N_l}$ is the ones dimension vector. The $G_{c,f}^{(l)}$ is a polynomial of $\hat{A}$ which is normalized adjacency matrix of graph as $\hat{A}=D^{-1/2}AD^{-1/2}$. Therefore,
\begin{equation}
	\begin{split}
		G_{c,f}^{(l)}=\sum_{k=0}^{K} g_{c,f,k}^{(l)}A^{k}
	\end{split}
\end{equation}
Here, $g_{c,f,k}^{(l)}$ is the graph filter polynomial coefficients. Thus, graph convolution operation becomes,
$\mathbf{x}_{f}^{(l+1)}=\sigma \left(\mathbf{y}_{f}^{(l)} \right)$, and $\sigma$ is ReLU activation function.
\par \textbf{GENeralized Aggregation Networks (GEN)} construct GCN network for mean-max aggregation functions as SoftMax and PowerMean \cite{gen2020}. The SoftMax aggregation is $\sum_{u\in \mathcal{N}(v)} \dfrac{exp(\beta m_{v,u})}{\sum_{i \in \mathcal{N}(v)}exp(\beta m_{v,i})}$, where, $m_{v,u}$ is the message set and $\beta$ is continuous variable. The PowerMean is expressed as $\left(\dfrac{1}{\mathcal{N}(v)} \sum_{u\in \mathcal{N}(v)} m_{v,u}^{p} \right)^{1/p}$, where $p$ is non-zero continuous variable. Therefore, GEN message construction is,
\begin{equation}
	\begin{split}
		m_{v,u}^{(l)}=\mathrm{ReLU}(h_{u}^{(l)}+\mathrm{I}(h_{e_{v,u}}^{(l)})\cdot h_{e_{v,u}}^{(l)})+\epsilon
	\end{split}
\end{equation}
where, $\mathrm{I}$ is the indicator function and $\epsilon$ is very small positive constant. Further, message aggregation function as $\zeta^{(l)}(\cdot)$ could be either SoftMax or PowerMean. This study extended to message normalization which combines other features during the vertex update phase. It can be expressed as,
\begin{equation}
	\begin{split}
		h_{v}^{(l+1)}=\mathrm{MLP}\left( h_{v}^{(l)}+s \cdot \|h_{v}^{(l)}\|_{2} \cdot \dfrac{m_{v}^{(l)}}{\|m_{v}^{(l)}\|_{2}} \right),
	\end{split}
\end{equation}
where $\mathrm{MLP}$ is the multi-layer perceptron, $s$ is the scaling factor and aggregated messages normalized to $\ell_{2}$ norm.

\section{GTAGCN}
\label{GTAGCN}
Our approach follows the conventional way of GNN working. We have adopted a message passing scheme that iteratively updates the representations of nodes. This allows to re-look at additional structural information to refine training stages. The proposed method extracts common properties of GEN \cite{gen2020} and TAGCN \cite{tagcn2018}, thus referred to as Generalized Topology Adaptive Graph Convolutional Networks (GTAGCN). The GTAGCN working can be expressed as,
\begin{equation}
	\label{GTAGCN}
	\begin{split}
		H^{(l)}=\mathrm{MLP}\left( \sum_{k=0}^{K} \mathrm{ReLU} \left(\hat{A}^{k} H^{(l-1)}W^{l}+\epsilon \right) \right)
	\end{split}
\end{equation}
Here, $\mathrm{MLP}$, $\mathrm{ReLU}$ and $\epsilon$ adopted from GEN. The iterative working of $\hat{A}^{k}$ for $K$ filters taken from TAG. The $\mathrm{MLP}$ is a deterministic process for efficient computations to train the network. This helps in adjusting the weights and enables the network to learn optimal weights that leads to meaningful representations. The $\mathrm{ReLU}$ non-linear property helps to train network systematically. This also results in complex relationships learning of data. The $\epsilon$ is a very small constant as discussed for GEN, which helps to retain value as non-zero. As discussed, $\hat{A}^{k}$ is $(D^{-1/2}AD^{-1/2})^k$, which refers to diagonal degree matrix. As discussed in \cite{Kipf2016}, the $D^{-1/2}AD^{-1/2}$ has eigenvalues and respond effectively with softmax function. Further, \cite{tagcn2018} defined recursive behaviour of $\hat{A}^{k}$ with respect to filters. This moves convolutional layers to go deeper and the output of the last layer is the projection of the first convolution layer. The use of filters increases representation capability in this process. The GTAGCN is applicable to both directed and undirected graphs as TAGCN works. The GNNs can exploit graph structures and recover hidden features which include useful information of graphs \cite{satoicml23}.      
\par The overall working of GTAGCN in GNN can be explained in sequential steps. The first step is initialization which assigns node or edge features to graph and prepare graph $G$ for the propagation. The propagation iterates over a fixed number of layers and a number of layers decided subject to the nature of data or adopted deep layered architecture. Here, each layer update node representation by aggregating information from neighboring nodes or edges. This step has been explained in equation \ref{GTAGCN}. The aggregated information updates node representation using a non-linear activation function. A readout function is applied after all layers are processed and results in graph level representation based on node level representations. This could be used at graph level classification. The proposed GTAGCN is applicable to both node and graph level classification. In addition, backpropagation helps to update model parameters based on difference between predicted and true labels. Recent studies suggest that GNNs are limited in their propagation operators and can be improved by incorporating trainable channel-wise weighting factors \cite{eliasoficml23}. Further, alternately optimized GNNs are other possibilities for semi-supervised learning on graphs from multi-view learning perspective \cite{hanicml23}. 
\par The message passing is a time-consuming step as it includes aggregation and combination. Also, node representation and intermediate matrices during forward or backward passes are storage consuming. Both time and space are common challenges in deep learning problems, so part of GTAGCN as GNN technique. The GTAGCN complexity is close to TAGCN in view to see the inherited nature of GTAGCN from TAGCN \cite{tagcn2018}. Further, dealing with sparse matrices as happen in many real-life datasets, reduces both time and space complexities in practice. The empirical performance of the proposed algorithm GTAGCN has been demonstrated in the next section \ref{experiments}. The GTAGCN update process employed to enhance scalability that could easily accommodate graph of varying sizes. This generalizes to different graph structures and its interpretability in model decision. The $\mathrm{MLP}$ is an established part of deep learning working, which could easily work for complex input values. The topology aware GNNs and inference-efficient $\mathrm{MLP}$ gaps are bridged using different distillation speeds and differential distributed graphs \cite{wuicml23}. As GNN enjoys the change of relationships over time for dynamic graphs, the use of $\epsilon$ avoids any zero situation. The overall combination is effective for diverse types of entities and complex system relationships. Attention mechanism applicable to relevant nodes as GTAGCN works close to techniques as GEN \cite{gen2020} and TAGCN \cite{tagcn2018}. 

\section{Experiments}
\label{experiments}
This section includes details for the datasets used, experiments setup and evaluation. We have covered sequence nature dataset as online handwriting recognition data \cite{ijdar2017}. Some of the datasets are directly available from repositories in the desired GNN form, other datasets are converted to GNN dataset form.

\subsection{Datasets}
The proposed GTAGCN technique has been implemented for GNN datasets such as cora, pubmed, citeseer, mnist, unipen and Gurmukhi HandWritten Text (GHWT)  datasets as presented in table \ref{tab1}. The cora, pubmed and citeseer are common datasets for GNN and are directly available in GNN form. The mnist in our experiments has been taken from the original mnist repository where 60k and 10k refer to train and test part of data. It is worth mentioning that we have not included superpixel based mnist dataset \cite{Monti2017}. The datasets as unipen and GHWT are handwritten strokes data for digits and Indic scripts Gurmukhi strokes. Therefore, we have converted mnist, unipen and indic to GNN dataset form. The datasets such as cora, pubmed and citeseer are node classification based data. The other datasets as mnist, unipen and GHWT are graph classification data. In mnist, each record is a graph and results in 60k train graphs and 10k test graphs. Similarly, for unipen and indic GNN graphs as depicted in table \ref{tab1}.
\begin{table*}
	\centering
	\caption{\textbf{GNN Datasets}} 
	\label{tab1}
	\begin{tabular}{ cccccc } 
		\hline
		\textbf{Dataset} & \textbf{graphs} & \textbf{nodes} & \textbf{edges} & \textbf{features} & \textbf{classes}\\
		\hline 
		citeseer \cite{citeseer} & 1 & 3327 & 9104 & 3703 & 6 \\
		cora \cite{cora} & 1 & 2708 & 10556 & 1433 & 7 \\
		Pubmed \cite{pubmed} & 1 & 19717 & 88648 & 500 & 3 \\
		mnist \cite{mnist} & 70000 & 31 & 30 & 31 & 10 \\
		unipen \cite{unipen} & 12154 & 25 & 24 & 25 & 10 \\
		GHWT \cite{ijdar2017} & 33215 & 25 & 24 & 25 & 62 \\
		\hline
	\end{tabular}
\end{table*}

\subsection{Experiment setup}
In all experiments, we use 3-layer $\mathrm{MLP}$ with batch normalization to map initial node representations to the desired dimensions, and 3-layer $\mathrm{MLP}$ without batch normalization for prediction. We have evaluated datasets using 10-cross validation using their provided data splits. The hidden dimension chosen size is 16. We apply dropout to the input of LSTM layer and chosen from $\{0,0,5\}$. We have chosen filter value $K=6$ to analyze the effect and performance of GTAGCN. The batch size is chosen as 64 for all datasets because of memory constraints. The Aadam optimizer is used with a fixed learning rate of $0.01$. The train and test split dataset has been implemented except for mnist dataset. The mnist dataset comes with 60k train and 10k test images. Therefore, we have followed the same standards. The unipen and GHWT datasets are split into $70:30$ as train and test parts. The datasets as cora, pubmed and citeseer train and test ratio have been taken as mentioned in their repository \cite{planetoid}. Further, we have trained model such that validation accuracy does not increase for 100 epochs. In order to use mnist, unipen and indic datasets, their GNN representation has been done. We have used chain code feature representations for each record of these three datasets as discussed in literature \cite{mtap2021, vjcs2015}. Further, these feature form based data converted to GNN datasets. 
\subsection{Results}
The tables \ref{tab2} and \ref{tab3} illustrate the classification accuracy achieved by the proposed GTAGCN and baselines on the six datasets mentioned in table \ref{tab1}. In view to see large number of GNN methods today \cite{mtap2023}, it is complex task to run more than 50 GNN algorithms. Especially, our study is more inclined to see the working of the hybrid technique preliminarily. Therefore, we have restricted selected GNN techniques in the following tables \ref{tab2} and \ref{tab3}. The results for all other techniques have been taken from the literature to avoid any confusion. We observe that our GTAGCN perform subject to state-of-the-art results for these datasets. We have been able to outperform results for two datasets as unipen and GHWT. In table \ref{tab2}, we find that our method is close to the results of other techniques. The datasets as cors, pubmed and citeseer in table \ref{tab2} are for node classification. The table \ref{tab3} demonstrates results for graph classification datasets as mnist, unipen and GHWT. We notice that GTAGCN outperforms for unipen and GHWT datasets. 
\par As results depicted in table \ref{tab2} for node classification datasets, we notice that GTAGCN perform at par with other popular techniques. Our results are very close to the best results for these datasets. Similarly, for graph classification results, we have achieved $99.10\%$, $99.16\%$ and $92.97\%$ recognition accuracy for mnist, unipen and GHWT datasets. The mnist best accuracy as $99.81\%$ has been noticed using support vector machine technique as discussed in literature \cite{suen2012}. The GNN based Chebyshev for mnist dataset accuracy is $99.14\%$ \cite{chebnet2016}. For unipen, \cite{ratz2003} have achieved $98.9\%$ accuracy, whereas we have been able to get $99.16\%$ accuracy. The GHWT accuracy in literature is $86.60\%$ and $87.71\%$ for train:test split as $70:30$ and $90:10$. We have achieved $92.97\%$ accuracy with $70:30$ as train:test split.

\begin{table*}[t]
	\caption{Classification accuracies (\%) for cora, Pubmed and Citeseer dataset}
	\label{tab2}
	\vskip 0.15in
	\begin{center}
		\begin{small}
			\begin{sc}
				\begin{tabular}{lccccr}
					\toprule
					Method & cora & Pubmed & Citeseer \\
					\midrule
					GCN \cite{Kipf2016}    & 81.5 & 79.0 & 70.3\\
					TAGCN \cite{tagcn2018}    & 83.3$\pm$ 0.7 & 81.1$\pm$ 0.4 & 71.4$\pm$ 0.5\\
					GAT \cite{gat2018}   & 83.0$\pm$ 0.7 & 79.0$\pm$ 0.3 & 72.5$\pm$ 0.7\\
					DCNN \cite{Monti2017}    & 76.8$\pm$ 0.6 & 73.0$\pm$ 0.5 & -\\
					MoNet \cite{Monti2017}  & 81.69$\pm$ 0.5 & 78.81$\pm$ 0.4 & -\\
					Chebyshev \cite{Kipf2016}     & 79.5 & 74.4 & 69.8\\
					GTAGCN (this paper)      & 82.2$\pm$ 0.5 & 79.1$\pm$ 0.4 & 70.1$\pm$ 0.5\\
					\bottomrule
				\end{tabular}
			\end{sc}
		\end{small}
	\end{center}
	\vskip -0.1in
\end{table*}

\begin{table}[t]
	\caption{Classification accuracies (\%) of GTAGCN for mnist, unipen and Indic dataset}
	\label{tab3}
	\vskip 0.15in
	\begin{center}
		\begin{small}
			\begin{sc}
				\begin{tabular}{lcccr}
					\toprule
					Dataset & Accuracy (\%) \\
					\midrule
					mnist & 99.10$\pm$ 0.7\\
					unipen & 99.16$\pm$ 0.5\\
					GHWT & 92.97$\pm$ 0.3\\
					\bottomrule
				\end{tabular}
			\end{sc}
		\end{small}
	\end{center}
	\vskip -0.1in
\end{table}

\section{General Observations}
As GNN research is in progression and further needs exploration of improvements in classification accuracy and performances. Some of the observations are listed here.
\begin{itemize}
    \item The GNN needs data in graph structural form. This includes a systematic understanding of data in terms of nodes, edges and respective relationships. 
    \item GNNs use message passing process for organizing graphs in a certain form that can be understood by machine learning algorithms. In it, every node is embedded with data about the node's location and its neighboring nodes. An AI model can find patterns and make predictions based on the embedded data.
    \item The GNN's ability to capture local and global features results in effective representation learning. However, data suitability for GNN working remains uncertain prior to experimentation.
    \item The GNN operator scalability feature could be generalized to large graphs which can efficiently process graphs of varying sizes and handle sparsity of graph structures.
    \item The GNN operator's nature could be in variance to graph permutations that helps to handle graph data regardless of any ordering.
    \item The GNN adaptability to new data needs transfer ability and acceptance of features at different stages of graph networks.
    \item The challenge of interpretability of learned representations for graphs is complex and its non-linear nature makes it difficult to interpret specific factors. This adds to the inner workings of GNN complex in nature and exhaustive, but the Mathematical nature of GNN overcomes these challenges.
    \item  The GNN operators include different propagation rules for updating node or edge representations. Also, differ in their respective aggregation mechanism for neighboring nodes. Also, respective network designs differ for message passing rules.
    \item GNNs are broadly classified as graph convolutional networks, recurrent graph neural networks, spatial graph convolutional networks, spectral graph convolutional networks and graph autoencoder networks.
    \item The GNN may face challenges for noisy or perturbed graphs. Interestingly, graphs quality to learn representations itself work effectively on noisy data.
    \item The GNN overall outcome depends on theoretical and its practical implementations as theoretical foundation helps in designing complex relationships and understanding of data properties.
    \item The various variants of GNNs as graph convolutional network (GCN), graph recurrent network (GRN) and graph attention network (GAT) have presented great performances on several deep learning tasks.
    \item While deploying GNN, we should care about model interpretability for building credibility, debugging or scientific discovery. The graph concepts that we should care, may vary from context to context. For example, with molecules we might care about the presence or absence of particular subgraphs    
    \item The GNNs have great applications in structural as well as non - structural scenarios. The applications of GNNs can be explored as graph generation, graph mining, graph clustering,  knowledge graphs, modeling real-world physical systems, molecular fingerprints, chemical reaction prediction, protein interface prediction, biomedical engineering, combinatorial optimization, traffic networks, image classification, visual reasoning and semantic segmentation etc.  
    \item  The GNN could be applied to various tasks based on texts. It could be applied for sentence level tasks such as text classification and word-level tasks such as sequence labeling. Further, it can be applied for other text based tasks as neural machine translation, semantic relation extraction between entities in texts, relational reasoning, event extraction, question answering and fact verification etc.
    \item One of the objective of present days GNN research is not making new GNN models and architectures only, but it is mainly about “constructing graphs”, more precisely, imbuing graphs with additional relations or structure that can be leveraged. So loosely it could seen as, the more graph attributes communicate the more we tend to have better GNN models.
    \item The GNN limitations do have a role in system performances. Common limitations are over-smoothing, computationally expensive nature for very large graphs, sensitive to perturbations in the graph structure, changes in graph structure during training, sparsity or irregularity of data.
    \item Overall, the graphs are a powerful and rich structured data type that have strengths and challenges. So we have also outlined some of the milestones that researchers have come up with in building GNN based models that process graphs. The success of GNNs in recent years creates a great opportunity for a wide range of new problems, and it is excited to see what the future researchers will bring.
\end{itemize}

\section{Conclusion}
We presented GTAGCN method which has been derived from two established techniques. We have been able to implement the proposed method to both sequence and static nature of data. The handwritten strokes in online form have been taken as sequenced data and static data as mnist images converted to a sequenced form of data. Further, established GNN datasets as core, pubmed and citeseer are experimented with the proposed GTAGCN method. The results are at par with the literature work and better in the case of sequenced data. We find that extraction of different techniques properties could be helpful to other domains' data where graph representation learning has not been explored yet.

\backmatter

\section*{Competing Interests}
The authors declare that they have no competing interests.






\bibliography{gentag_bib}

\end{document}